\documentclass{article}
\usepackage{graphicx} 
\usepackage{amsmath}
\usepackage{amssymb}
\usepackage{natbib}

\title{ImplicitHop: Evaluating knowledge integration across context and memory}

\author{%
Alexander Detkov, Dashiel Karish,\\
Matthew Oleynikov, Matt Thomson\\[0.5em]
\normalsize California Institute of Technology\\[0.25em]
\small All authors contributed equally.
}
\date{October 2026}

\begin{document}

\maketitle

\begin{abstract}
The ability to integrate new information and deduce its consequences is central to learning, understanding, and reasoning in large language models (LLMs). Whether information arrives during training or in-context, models must not only store local facts, but propagate their long-range implications globally. To explore information integration in LLMs, we introduce two complementary natural language benchmarks for large-scale evaluation of deductive reasoning as a function of inference depth. The first is a formal graph benchmark enabling controlled evaluation over a wide range of graph topologies and integration tasks; the second is a knowledge composition benchmark grounded in real-world knowledge graphs. We use these benchmarks to compare knowledge integration from information encoded in context versus information encoded in model weights during pretraining, evaluating a set of open models implicitly, without chain of thought. We find that accuracy decays sharply with increased composition length for both pretrained and in-context knowledge. Information presented in-context, however, substantially outperforms deduction over information encoded in weights. These results quantify a fundamental limitation of current models with direct implications for prompt-based control of agent behavior and for continual learning strategies, and they provide a foundation for developing training, prompting, and integration methods that support deeper reasoning.

\end{abstract}

\section{Introduction}
AI systems receive information from pretraining, prompts, tool outputs, and weight updates. To reason and act effectively, knowledge must not only be stored, but combined to infer new information. Models often fail at integrating information, even when they know the necessary facts. A model that learns that X has been elected president may recall the new fact, yet fail to answer ``Who is the president's spouse?'' even though it knows that X is married to Y \citep{zhong2023mquake, cohen2024ripple}. Models that answer two questions correctly also often fail the question that composes them \citep{press2023compositionality}.

To measure knowledge integration we create two benchmarks over varied tasks that quantify integration by inference depth, measured in hops: the number of facts that must be integrated to reach an answer. In the first, a random graph is given in the prompt as a list of edges, and the model answers questions about the nodes exactly \(k\) hops from a start node. In the second, the model answers questions that compose \(k\) real Wikidata facts, such as ``In what year was the director of Jaws born?'' at \(k=2\). The facts are given in the prompt, stored only in the model's weights, or both, so we can compare how well models integrate knowledge from context and from memory.

Across 15 open models with reasoning off, accuracy falls sharply with hops. On graphs, models name a neighbor of the start node \(99\)--\(100\%\) of the time, but a node two hops away only \(9\)--\(40\%\) of the time. On Wikidata, models compose facts given in the prompt at \(\geq 94\%\) up to \(k=3\), but facts in memory at only \(37\)--\(64\%\) at \(k=2\) and \(\leq 12\%\) from \(k=4\), although every fact is known. Models integrate information in context far better than information in their weights, and both only to shallow depth.

\section{Datasets}

\subsection{Synthetic Graph}
\paragraph{Problem instances:}A problem instance is defined by a finite, simple, undirected graph $G=(V,E)$ with a vertex set $V$, $|V|=n$, and edge set $E\subseteq \binom{V}{2}$, $|E|=m$, together with a question $q$ about $G$ and its answer $y=f_q(G)$. Note that the mean degree is $\bar d=2m/n$. A \emph{hop} is the traversal across a single edge in $G$. The \emph{graph distance} $\operatorname{dist}_G(u,v)$ is the minimum number of hops on any path from $u$ to $v$, with $\operatorname{dist}_G(u,u)=0$ and $\operatorname{dist}_G(u,v)=\infty$ when no path exists. For an \emph{anchor} vertex $a\in V$ and an integer $k\ge 0$, the \emph{exactly-k-hop shell} of $a$ is
\begin{equation}
    S_k(a)=\{u\in V:\operatorname{dist}_G(a,u)=k\}
\end{equation}
For example, $S_1(a)$ is the neighborhood of $a$, and $S_2(a)$ contains the vertices reachable in two hops but not in fewer hops. Being exactly $k$ hops from $a$ refers to only the \emph{shortest} path. A vertex that is adjacent to $a$ and can also be reached by a two-hop path from $a$ belongs in $S_1(a)$ and not $S_2(a)$. Every vertex outside of the connected component of $a$ has distance $\infty$ and lies in no shell.

\paragraph{Random graph model:}All graphs are generated using the $G(n,p)$ Erdős--\allowbreak Rényi model, in which each of the $\binom{n}{2}$ possible edges is included independently with probability $p$. Rather than fixing $p$, we fix a target expected mean degree $d_\star$ and set
\begin{equation}
    p=\frac{d_\star}{n-1},\quad \text{so that} \quad \mathbb{E}[\deg(v)]=(n-1)p=d_\star,
\end{equation}
which keeps the expected degree fixed across graph sizes. For each graph, $n$ is drawn uniformly from an integer range $[n_{min},n_{max}]$, and $p$ is then set from the drawn $n$. Vertices are given random labels: $n$ distinct two-character strings are generated from A through Z without I and O. Random labels ensure that neither the label order nor the label identity carries information about the structure. Each graph is generated from its own seed, derived from a global seed (fixed at $0$) and its index in the graph pool, so every graph can be reproduced.

Connectivity is not a condition that is required. At the target expected mean degree $d_\star=3$, graphs need not be connected, and conditioning on connectivity would change the sampling distribution. Empty shells and unreachable vertices are part of the benchmark and have defined answers.

\paragraph{Sub-tasks:} On each graph we have four operations at each hop level $k\in\{0,1,2,3\}$, giving a $4 \times 4$ grid of sub-tasks $T_{\text{op},k}$. For $k\ge 1$, every operation is posed over the shell $S_k(a)$ of an anchor $a$:
\begin{align*}
    \text{verify} &: (a,b) \mapsto \begin{cases}
        1,\ b\in S_k(a)\\
        0,\ \text{else}
    \end{cases} && \parbox{0.48\textwidth}{``Is $b$ exactly $k$ hops from $a$?''}\\
    \text{count} &:  a \mapsto |S_k(a)| && \parbox{0.48\textwidth}{``How many vertices are exactly $k$ hops from $a$?''}\\ 
    \text{name\_one} &: a \mapsto \text{any } v\in S_k(a) && \parbox{0.48\textwidth}{``Name one vertex exactly $k$ hops from $a$.''}\\ 
    \text{enumerate} &: a \mapsto S_k(a) && \parbox{0.48\textwidth}{``List every vertex exactly $k$ hops from $a$.''}
\end{align*}
Each column in the grid (one operation, varying $k$) would then isolate the scale, and each row (one $k$, varying operation) would then isolate the level of output, ranging from one bit (verify) to an exact set (enumerate). The level $k=0$ requires a different approach because $S_0(a)=\{a\}$, making all four questions trivial. At $k=0$ the four operations are posed against the whole vertex set: whether a given label is in $V$, $|V|$, any element of $V$, and all of $V$.

\paragraph{Natural language representation:} Each problem instance must be turned into text before it can be given to a language model. The prompt states that the graph is undirected, lists the vertices by their labels, then lists the edges (Edges: AB-CD, YE-LY, \dots), and ends with the question.

\subsection{Wikidata Graphs}
A knowledge graph \((V,R,E)\) is a set of entities \(V\), a set of relations \(R\), and for each relation
\(r\in R\) a set of directed edges \(E_r\subseteq V\times V\), where \((u,v)\in E_r\) states that \(v\) is
the \(r\) of \(u\). A \(k\)-hop chain is a path \(c=(v_0, r_1, v_1, \dots, r_k, v_k)\) composed of local \(1\)-hops \((v_{i-1}, r_i, v_{i})\)
with \((v_{i-1}, v_i)\in E_{r_i}\) for \(i=1,\dots,k\).

We build the graph from Wikidata, using 30 single-valued relations (e.g.\ director,
founder, father, birth year) between popular entities. We sample chains of \(k=1,\dots,6\) hops by
random walks, removing any chain that can be answered without composing: no two non-consecutive
entities may be directly linked in Wikidata, and no entity may appear in the name of an earlier one.
We keep 200 chains per length \(k\).

For each \(k\)-chain, the required \(k\) one-hop facts are retrieved from model weights, provided in context, or both. If the facts are in the weights, we ensure that the model correctly answers each 1-hop fact posed as a question; otherwise, the \(k\)-chain is not used in the model's evaluation. For the in-context condition, the chain's facts
are stated in the prompt with every entity given an invented name, so the model cannot rely on memory. Sixteen decoy chains with the same relations are mixed in, so the model must follow the chain from its start and cannot take semantic shortcuts.

\section{Results}
\subsection{Synthetic Graphs}

\begin{figure*}[t]
    \centering
    \includegraphics[width=\textwidth]{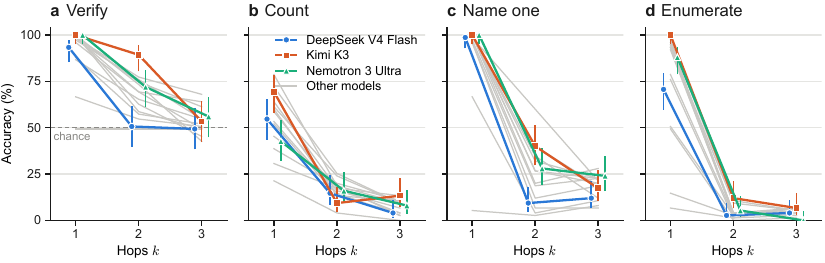}
    \caption{\textbf{Models receive graph adjacency in-context but integrate the information poorly, failing at long-range inferences.}
DeepSeek, Kimi, Nemotron, and 12 other models answer questions about a random graph of 50--60 nodes given in the prompt as a list of edges, with reasoning off. Each question concerns the nodes exactly \(k\) hops from a start node, so \(k=1\) questions require only direct-neighbor information.
\textbf{a)} Verify: is a given node exactly \(k\) hops away? Accuracy falls from \(93\)--\(100\%\) at \(k=1\) to \(49\)--\(56\%\) at \(k=3\), close to chance.
\textbf{b)} Count: how many nodes are exactly \(k\) hops away? Models count correctly \(43\)--\(69\%\) of the time at \(k=1\) and \(\leq 16\%\) from \(k=2\).
\textbf{c)} Name one: name a node exactly \(k\) hops away. Models name a neighbor at \(99\)--\(100\%\), but a node two hops away at only \(9\)--\(40\%\).
\textbf{d)} Enumerate: list every node exactly \(k\) hops away. Models list all neighbors at \(71\)--\(100\%\), but no model lists more than \(12\%\) of sets correctly from \(k=2\).}
    \label{fig:graphbench}
\end{figure*}

\subsubsection{Accuracy collapses at two hops}
Accuracy collapses when a question requires \(\geq2\)-hop reasoning. DeepSeek, Kimi, and Nemotron name a neighbor of the start node \(99\)--\(100\%\) of the time, but a node two hops away only \(9\)--\(40\%\) of the time and a node three hops away only \(12\)--\(24\%\) (Figure \ref{fig:graphbench}c). The same drop occurs across tasks: from \(k=2\), counting falls to \(\leq 16\%\) and enumerating to \(\leq 12\%\), and verifying is near chance by \(k=3\) (Figure \ref{fig:graphbench}a,b,d). The other 12 models, from 0.8B to 2.8T parameters, follow the same pattern. 

\subsection{Wikidata Graphs}

\begin{figure*}[t]
    \centering
    \includegraphics[width=\textwidth]{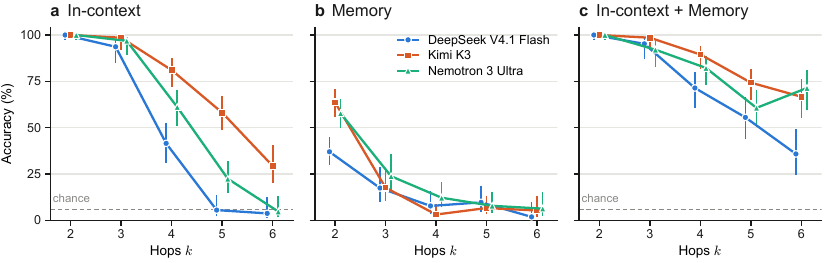}
    \caption{\textbf{Models compose facts from the prompt far better than from memory.} 
DeepSeek, Kimi, and Nemotron answer \(k\)-hop Wikidata questions with reasoning off and information given in context, in memory, or both. \textbf{a)} The chain's facts are given in the prompt under invented names, among 16 decoy chains with the same relations. All models compose at \(\geq 94\%\) up to \(k=3\), but at \(k=6\), Kimi composes at \(29\%\), while DeepSeek and Nemotron
Ultra fall to chance. \textbf{b)} The facts are recalled from memory without supporting facts in the prompt. Composition
reaches only \(37\)--\(64\%\) at \(k=2\) and \(\leq 12\%\) from \(k=4\), although every fact is known.
\textbf{c)} The facts are given in the prompt under their real names, among 16 real decoy chains, so the
model can use both the prompt and memory. Accuracy reaches \(36\)--\(71\%\) at \(k=6\).}
    \label{fig:wikidata}
\end{figure*}

\subsubsection{Composition accuracy decreases sharply with number of hops}
Composition accuracy decreases sharply as chains grow longer. When information is presented in context, Kimi composes \(99\%\) of chains at \(k=3\), \(81\%\) at \(k=4\), \(58\%\) at \(k=5\), and \(29\%\) at \(k=6\), while DeepSeek
and Nemotron are indistinguishable from chance by \(k=5\) and \(k=6\), respectively (Figure \ref{fig:wikidata}a). Similar accuracy decay with composition length occurs when information is in the weights or is available both in context and in memory. Even when every fact is known or given in the prompt, models struggle to compose information.

\subsubsection{Models compose facts in context but not from memory}
Models compose information given in context far better than information recalled from memory. At \(k=2\),
every model composes \(100\%\) of chains in context but only \(37\)--\(64\%\) from memory, and from \(k=4\)
memory composition falls to \(\leq 12\%\) (Figure \ref{fig:wikidata}b). Giving the facts in context under
their real names, so models can use both prompt and memory, yields the highest composition accuracy, with \(k=6\) accuracy reaching
\(36\)--\(71\%\), compared to \(4\)--\(29\%\) in context alone and \(2\)--\(6\%\) in memory alone (Figure \ref{fig:wikidata}c). These results suggest that models compose known facts reliably only when the information is in context.

\bibliographystyle{plainnat}
\bibliography{references}

\end{document}